\documentclass[final]{hld2024} 

\title[Probing LLMs with Cumulants of Softmax Entropy]{Probing Geometry of Next Token Prediction Using Cumulant Expansion of the Softmax Entropy}

\hldauthor{%
\Name{Karthik Viswanathan$^{*}$} 
\Email{k.viswanathan@uva.nl}\\
\addr {Institute of Physics, University of Amsterdam, the Netherlands} 
\AND
\Name{Sang Eon Park$^{*}$} \Email{sangeon@mit.edu}\\
\addr {Department of Physics, Massachusetts Institute of Technology, Cambridge, U.S.A.} }

\usepackage{amsmath}
\usepackage{amssymb}
\usepackage{mathtools}
\usepackage{thmtools}
\usepackage{xcolor}
\usepackage{comment}
\usepackage{booktabs}
\newcommand{\hide}[1]{} 

\DeclareMathOperator*{\argmin}{arg\,min}

\newenvironment{talign*}{\let\displaystyle\textstyle\csname align*\endcsname}{\endalign}
\definecolor{airforceblue}{rgb}{0.36,0.54,0.66}
\definecolor{alizarin}{rgb}{0.82, 0.1, 0.26}
\newcommand{\kv}[1]{\textcolor{airforceblue}{\textsf{(KV: #1)}}}
\newcommand{\sap}[1]{\textcolor{alizarin}{\textsf{(SP: #1)}}}
\begin{document}

\maketitle

\begin{abstract}%
We introduce a cumulant-expansion framework for quantifying how large language models (LLMs) internalize higher-order statistical structure during next-token prediction. By treating the softmax entropy of each layer’s logit distribution as a perturbation around its “center” distribution, we derive closed-form cumulant observables that isolate successively higher-order correlations. Empirically, we track these cumulants in GPT-2 and Pythia models on Pile-10K prompts. (i) Structured prompts exhibit a characteristic rise–and–plateau profile across layers, whereas token-shuffled prompts remain flat, revealing the dependence of the cumulant profile on meaningful context. (ii) During training, all cumulants increase monotonically before saturating, directly visualizing the model’s progression from capturing variance to learning skew, kurtosis, and higher-order statistical structures. (iii) Mathematical prompts show distinct cumulant signatures compared to general text, quantifying how models employ fundamentally different processing mechanisms for mathematical versus linguistic content. Together, these results establish cumulant analysis as a lightweight, mathematically grounded probe of feature-learning dynamics in high-dimensional neural networks.

\end{abstract}


\begingroup
  \def\thefootnote{$^{*}$}
  \makeatletter
  \long\def\@makefntext#1{\parindent 1em$^{*}$ #1}
  \makeatother
  \footnotetext{  Equal contribution}
\endgroup


\section{Introduction}

Growing evidence suggests that Large Language Models (LLMs) based on transformer architectures process information in distinct stages~\cite{lad2024remarkable, cheng2024emergenceofabstraction,  skean2025layer}. By tracking statistical quantities of latent prediction probabilities \cite{belrose2023eliciting} such as mean, KL-divergence, and entropy across layers, as well as geometric properties of internal representations like intrinsic dimension \cite{valeriani23} and curvature \cite{skean2025layer}, researchers can investigate the emergent representations within these models.

We introduce a novel method for examining information evolution within different LLM layers using the cumulant expansion of the entropy from softmax-generated probability distributions (hereafter referred to as 'softmax entropy'). In both training and inference of LLMs, the softmax function serves as a critical bridge translating between internal model representations and next-token prediction. By analyzing LLM through the lens of information theory, particularly through cumulants, we formalize how information is learned during each training step and how next token predictions are refined across layers. In this paper, we propose that cumulants of softmax entropy effectively capture the emergence of higher-order moments and demonstrate their efficacy in the study of off-the-shelf LLMs through i) experiments with structured and shuffled prompts, and ii) studying the evolution of cumulants during training. We believe that our new observable provides mathematical language for explicitly demonstrating neural network learning higher-order correlations within data.


\section{Related Work}


Several recent studies have investigated the evolution of LLM representations through their layers, examining information refinement and emergent behaviors within these architectures~\cite{lad2024remarkable, cheng2024emergenceofabstraction, skean2025layer}. Mean field theoretical approaches offer complementary insight into the dynamics in neural network~\cite{fischercorrelationfunctions, Segadlo_2022, szekely2024learning, Helias_2020, beyondmeanfield, mathematicalperspective, castin2024smoothattention}. Analyzing internal representations points to distributional simplicity bias, wherein neural networks learn lower-order statistics of input data before progressing to higher-order statistics ~\cite{refinettidistributionalsimplicitybias, belrosedistributionalsimplicitybias, rende2024a}. 

Information theory has emerged as a valuable tool for understanding different phenomena in large language models, as demonstrated in previous works~\cite{ton2024understandingchainofthoughtllmsinformation, dombrowski2024an, conklin2024representationslanguageinformationtheoreticframework, ali2025entropy}. The cumulant expansion, a well-established method in statistical physics, has found applications across diverse fields, including astrophysics~\cite{Boyle2023cumulant} and chemical physics~\cite{ensemble_refinement}. In particular,~\cite{cumulant_expansion_barber} expresses KL divergence as an expansion in cumulants for intractable distributions. Complementary to our approach,~\cite{viswanathan2025geometrytokensinternalrepresentations} establishes a relationship between intrinsic dimension and softmax entropy in internal representations, while~\cite{gurnee2024universalneuronsgpt2language} examines skew and kurtosis of token activations to understand universal neuron properties.

\section{Cumulants as Observables}
\label{sec: cumulants_as_observables}

\begin{figure}[t]
    \centering
    \includegraphics[width=0.7\textwidth]{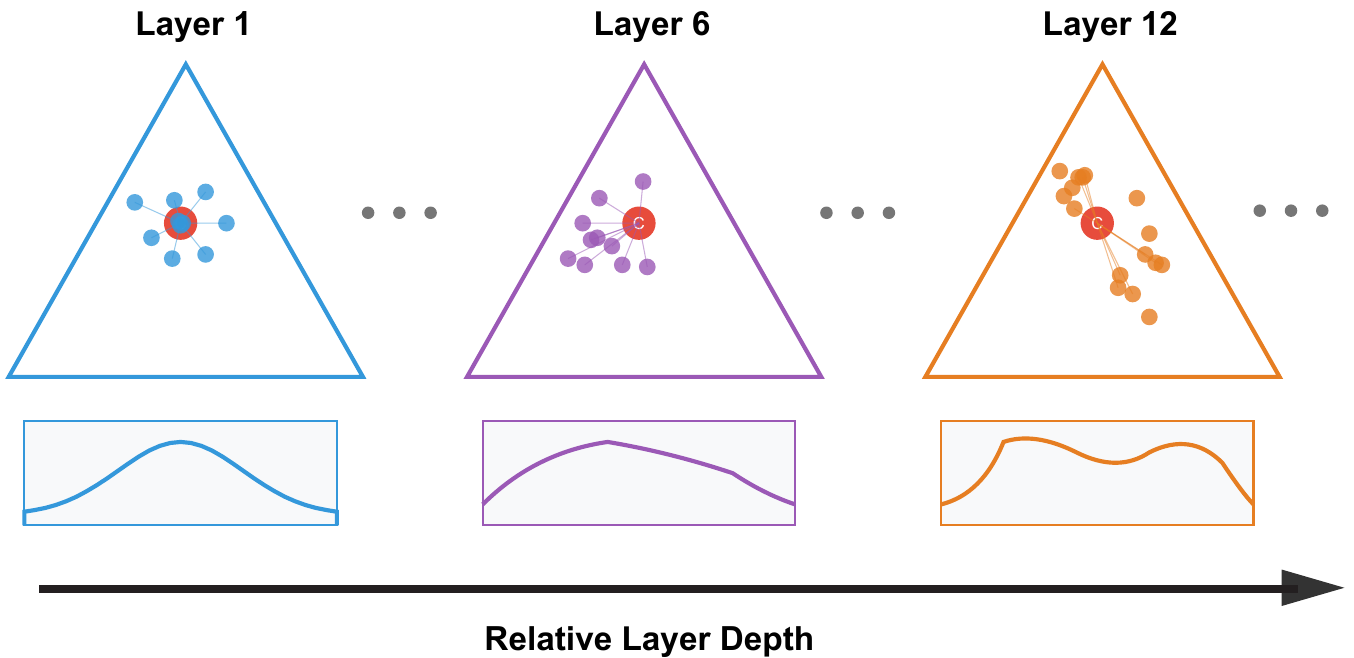}
    \caption{{\bfseries Schematic of Logit Geometry Across Layers.}  
    Each triangle represents the probability simplex at a given layer, where colored dots correspond to token logits mapped to probabilities. The red circle labeled ‘c’ indicates the probability of the center of logits. The histograms illustrate the distribution of token-wise deviations from the center.}
    \label{fig:schematicdiagram}
\end{figure}

We propose cumulants from the expansion of the softmax entropy as variables to probe the internal representation of the tokens. Let $\mathbf{X} = \{X_1, X_2, ... X_N\}$  be the collection of logits for $N$ tokens, and let $\boldsymbol{\mu}$ be the "center" of the logits defined as the point minimizing the sum of \textit{distances} to the sample points $\mathbf{X}$, where the \textit{distance} is the KL-divergence from the minimizing point to the sample points, 
\begin{equation}
\label{eq:centerdef}
\boldsymbol{\mu} = \argmin_{Y} \sum_{i} D_{KL}\left(p(X_i)||p(Y)\right) \quad \Rightarrow \quad p(\boldsymbol{\mu}) = \frac{1}{N} \sum_{i =1}^N p(x_i)
\end{equation}

Eq.~\ref{eq:centerdef} shows that the next token prediction corresponding to the center is the average of the next token prediction of the collection of logits.

Let $S(X_i)$ be the softmax entropy of the token $X_i$, defined as the entropy of the discrete distribution over the vocabulary after applying the softmax function. Let $\langle S(\mathbf{X}) \rangle$ be the arithmetic mean of the softmax entropy of all the tokens in a prompt. Then we can perform the perturbative expansion of $S(\mathbf{X})$ around the softmax entropy of the center $S(\boldsymbol{\mu})$.

\begin{equation}
\label{eq:softmaxaroundmean}
    \langle S(\mathbf{X}) \rangle = S(\boldsymbol{\mu}) - \frac{1}{N} \sum_{i = 1}^ND_{KL}(p_i||p_{\boldsymbol{\mu}})
\end{equation}
where $p_i = \operatorname{softmax}(X_i)$ denotes the softmax probability distribution over the vocabulary for token $i$, and $p_{\boldsymbol{\mu}}$ the softmax probability distribution for the center of logits, as defined in Eq.~\ref{eq:centerdef}. Using Eq.~\ref{eq:softmaxaroundmean},  we can motivate $ \frac{1}{N} \sum_{i = 1}^N D_{\mathrm{KL}}(p_i \,\|\, p_{\boldsymbol{\mu}})$ as a measure of interaction between logits and the center. The underlying intuition is that the entropy of the prompt is obtained by subtracting this interaction term from the entropy of the center. This view is supported by experimental results in Section~\ref{Sec:Results}. By expanding the KL-divergence, the second term in the sum of the right-hand side of Eq.~\ref{eq:softmaxaroundmean} can be expressed as cumulants of the softmax probability distribution of $\mathbf{X}$. 

\begin{equation}
\label{eq:kldivergenceascumulants}
    D_{\text{KL}}(p_\beta(X_i) \| p_\beta(\boldsymbol{\mu})) = \sum_{n = 2}^{\infty} \frac{\beta^n}{n!} \kappa^{p_\beta(X_i)}_n(-\delta X_i)
\end{equation}


Eq.~\ref{eq:kldivergenceascumulants} shows concretely how the KL divergence can be expressed in terms of cumulants using cumulant generating functions, where $\delta X_i$ is a random variable associated with the $i^{\text{th}}$ token in a prompt, constructed by sampling a vocabulary token $j$ from the softmax probability distribution $p_\beta(X_i)$ and assigning it the value $\delta X_{i j} = X_{ij} - \mu_{j}$. $\beta = \frac{1}{T}$, $T$ is the temperature. The superscript $p_{\beta}(X_i)$ denotes the underlying distribution of the random variables $X_i$ used in $\delta X_i$. 

Plugging this back into Eq.~\ref{eq:softmaxaroundmean}, we finally obtain the relation in Eq.~\ref{eq:finalrelation}. 

\begin{equation}
\label{eq:finalrelation}
     \langle S(\mathbf{X}) \rangle = S(\boldsymbol{\mu}) - \frac{1}{N} \sum_{n = 2}^{\infty}   \frac{\beta^n}{n!} \kappa^{p_\beta(\mathbf{X})}_n\left(-\sum_{i = 1}^N \delta X_i \right)
\end{equation}


Eq.~\ref{eq:finalrelation} demonstrates that by observing cumulants, we effectively measure the distribution of logits—specifically, how token logits are distributed relative to the center logit. Furthermore, this equation reveals that token-wise average cumulants correspond to the cumulants of the aggregated random variable $\delta X = \sum_{i = 1}^n{\delta X_i}$. We verify this with Monte-Carlo simulation in Appendix~\ref {subsec:verificationcumulants}.

This implies that prompts with different $\delta X$ distributions produce distinct cumulants; conversely, identical $\delta X$ distributions yield the same cumulants. In theory, the average cumulants contain the same information as the distribution $p_\beta(\delta X)$, and this equivalence offers insight into what our observables capture. However, accurately estimating $p_\beta(\delta X)$ directly from $\delta X_i$ requires many simulations. Hence, computing the average of token-wise cumulants provides a more computationally efficient and reliable way of probing $p_\beta(\delta X)$ in practice.

We could also make an analogy to the perturbative Taylor series expansion, where we are performing the perturbative expansion of the entropy, and by calculating higher-order cumulants, we consider successively higher-order correlations within token logits. 
The cumulant summary statistics we introduce probe the geometry of the latent probabilities that reside in the probability simplex. In particular, different sets of logits that yield the same set of probabilities post-softmax will produce identical cumulants.
We show the full derivation of each equation, and how we calculate each cumulant in practice in Appendix~\ref{sec:derivations} and ~\ref{sec:cumulantcalculation}. 

\section{Results}
\label{Sec:Results}

We use the Pile-10K dataset \citep{NeelNanda_pile-10k}, a subset of the Pile dataset \citep{pile}, as a representative sample of diverse textual data. We first show the result with the $3218^{\text{th}}$ prompt from the Pile-10K dataset, with the Pile set name: \textit{ArXiv}. We then compare the results for two different types of prompts: general text and mathematics prompts. We also show that the results generalize to other prompts in the dataset, as well as different models in Appendix~\ref{sec:suppfigs}.

\subsection{Cumulants in Structured and Shuffled Prompts}

We refer to the original prompts as "structured" prompts, while prompts with randomly permuted token orders are called "shuffled" prompts. To obtain latent predictions from the intermediate layers, we employ TunedLens~\cite{belrose2023eliciting}. For all results, We normalize the $n$th cumulant by dividing it by $n!$ to reflect its contribution to the softmax entropy as seen in Eq.~\ref{eq:finalrelation}.

\begin{figure}[t]
    \centering
    \includegraphics[width=1.0\textwidth]{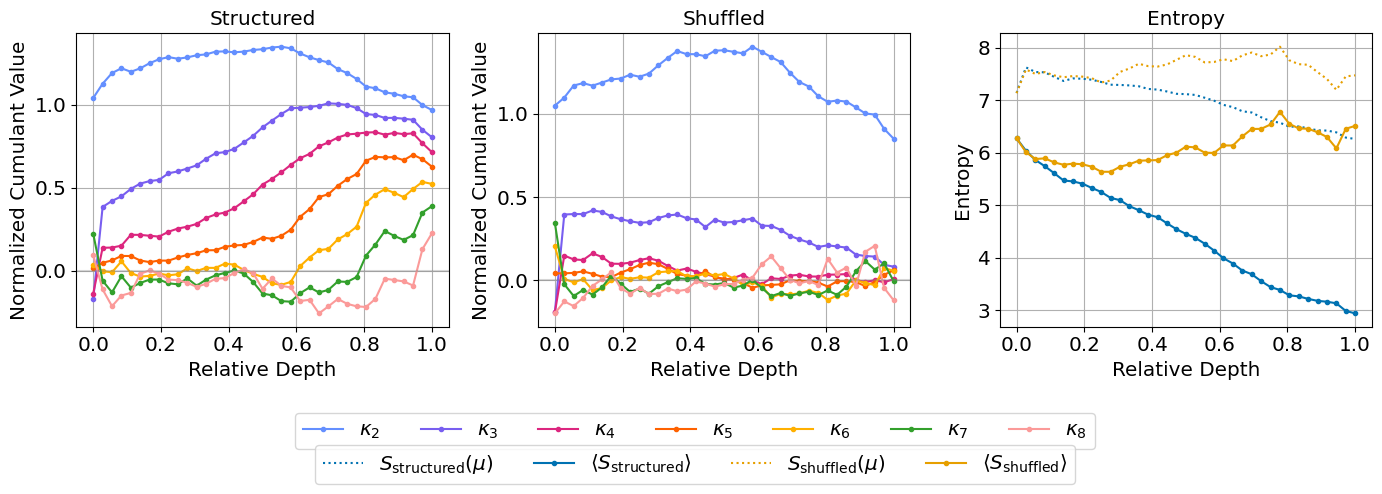}
    \caption{{\bfseries Cumulants in Structured and Shuffled Prompts.}  
    Left: Cumulants across layers for a single structured prompt (3218 from Pile 10K) in GPT-2 Large.
    Middle: Cumulants for the shuffled version of the same prompt.  
    Right: Comparing mean softmax entropy (solid lines) and the entropy of the center (dotted lines) for both structured and shuffled prompts. }
    \label{fig:structured_shuffled}
\end{figure}

We compare the cumulant profiles of a structured prompt (left) and its shuffled counterpart (middle), noting that the second cumulant remains consistently near 1.0 in both cases. The structured prompt exhibits a clear depth-dependent trend for the higher-order cumulants: they increase through intermediate layers, plateau, and then slightly decline near the final layers. 
In contrast, the cumulants for the shuffled prompt (middle plot) remain relatively constant around zero for higher cumulants, with significantly less variation than those of the structured prompt. We conclude that when the model captures patterns within data, token logit deviations from the center develop a characteristic pattern across layers.

A more striking result appears in the right panel, which compares the mean softmax entropy with the softmax entropy of the center. For structured prompts (blue lines), the gap between the center's softmax entropy (dotted line) and the mean softmax entropy of all tokens (solid line) increases with the layers. 
The result indicates stronger higher-order relations in deeper layers for structured prompts, with stronger interaction between the center and the logits. In contrast, for shuffled prompts (yellow lines), the higher-order cumulants remain mostly constant. 




To test for the dependency on prompts and models, we rerun the experiment on different prompts from the Pile dataset and different size models from the GPT-2 family. The pattern we observe is almost identical across all prompts and models, and can be seen in Appendix~\ref{subsec:otherprompts}.

\subsection{Evolution of Cumulants during Training}

\begin{figure}[t]
    \centering
    \includegraphics[width=0.8\textwidth]{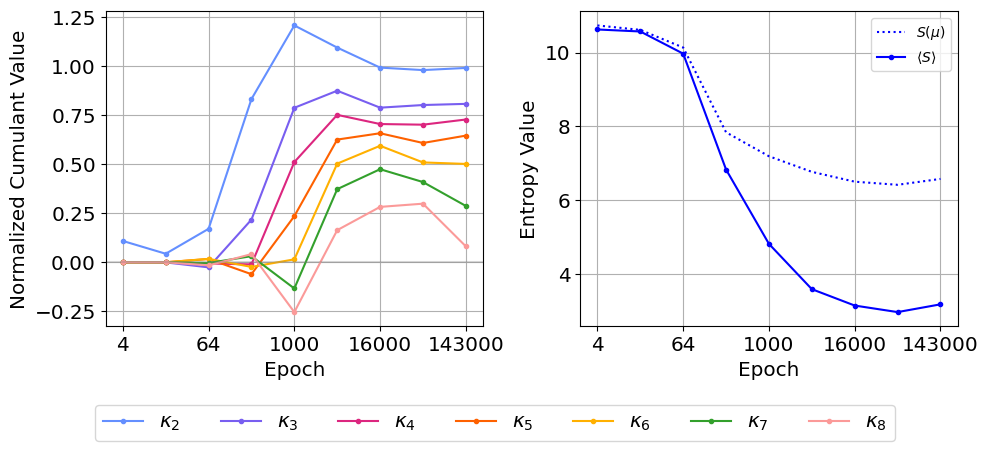}
    \caption{{\bfseries Evolution of Cumulants During Training.}  
    Left: Cumulants across layers of the Pythia-160M model tracked over training epochs.  
    Right: mean softmax entropy (solid line) and softmax entropy of the center (dotted line) as a function of training.  
    }
    \label{fig:training}
\end{figure}

We further investigate the evolution of cumulants during training and what they reveal about neural network learning dynamics. The left panel of Fig~\ref{fig:training} shows that all cumulants increase before plateauing as training progresses. This indicates that the logit distribution—specifically, the distribution of deviations from the center—becomes increasingly complex. We observe larger spread (variance), greater asymmetry (skewness, related to the third moment), heavier tails (kurtosis, from the fourth moment), and other higher-order statistical properties.

The right panel of Fig~\ref{fig:training} demonstrates that as training progresses, the discrepancy between the center's softmax entropy and the mean softmax entropy of all token logits increases. This growing gap indicates that the center becomes less capable of explaining the total entropy, requiring higher-order information to accurately characterize the logit distribution. This pattern explicitly demonstrates that models progressively learn higher-order correlations within data during training.

\subsection{DM Mathematics vs Pile-CC: A Cumulant Perspective}
\begin{figure}[t]
    \centering
    \includegraphics[scale = 0.3]{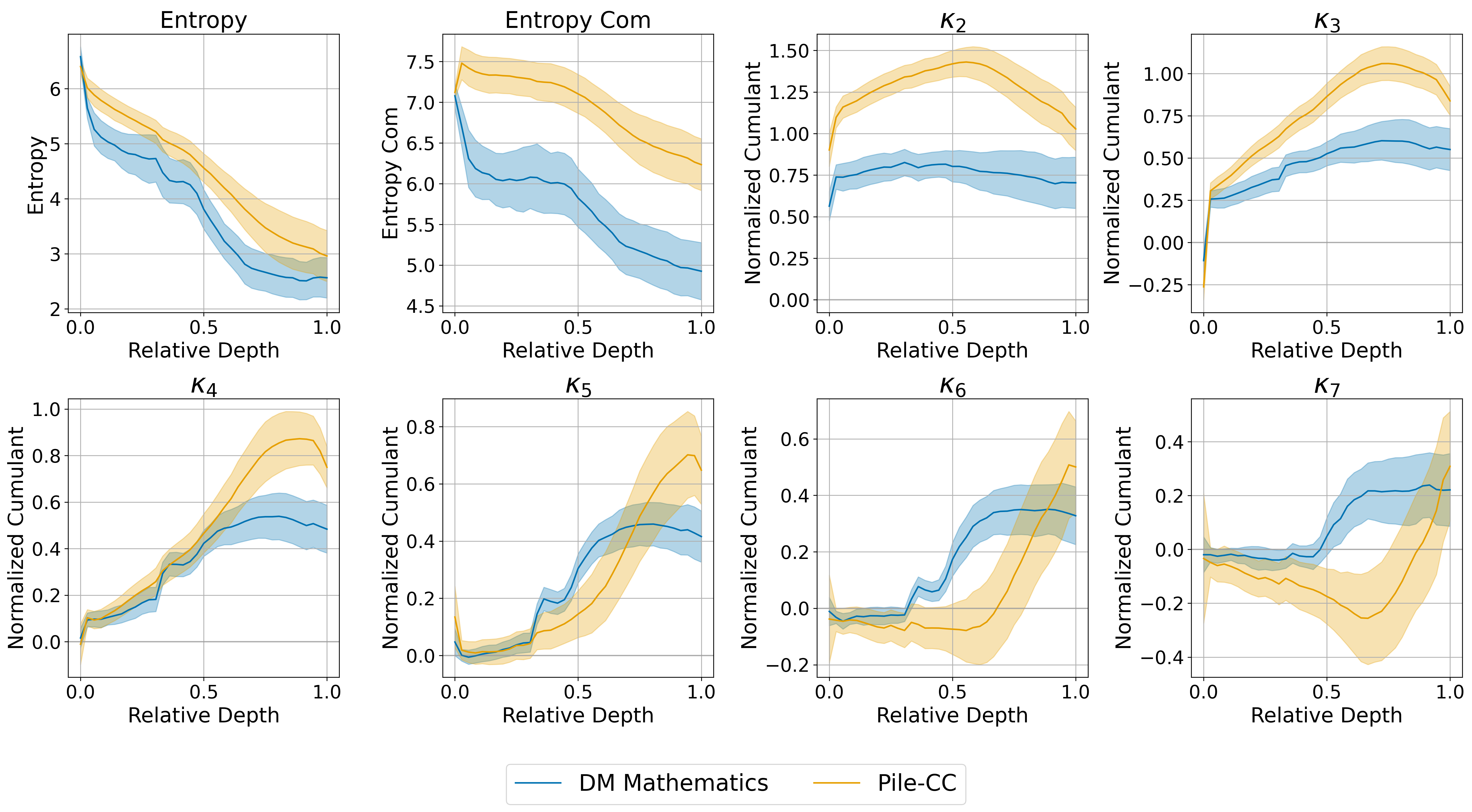}
    \caption{{\bfseries Cumulants in DM Mathematics and Pile-CC Prompts.}  
    Comparison of normalized cumulants ($\kappa_2$ through $\kappa_7$) and entropy measures across model layers for mathematical prompts (DM Mathematics topic with $99$ prompts) versus general web text (Pile-CC topic with $570$ prompts) in GPT-2 Large. Each plot shows the mean (solid line) and standard deviation (shaded region) computed across multiple prompts from each dataset. Mathematical prompts exhibit distinct cumulant profiles compared to general text.}
    \label{fig:cumulant_comparison_topics}
\end{figure}

Prior research \cite{Li_2025, wang2025differentiation} has established that contexts involving logic and language produce distinct geometric properties in neural network representations. To compare how code/math prompts are processed differently from general text processing in language models, we analyze the cumulant-based statistical properties of next-token prediction as it evolves across layers. We compare prompts from DM Mathematics against Pile-CC, representing general web text from the Pile-10k \cite{NeelNanda_pile-10k} dataset in Figure \ref{fig:cumulant_comparison_topics}, where we notice distinct statistical signatures between mathematical and general text processing through cumulant analysis. 
Mathematical prompts show lower entropy than Pile-CC prompts across layers, though the separability between distributions remains modest. We observe clearer differences in the entropy of the center $S(\mu)$ and lower-order cumulants $\kappa_2$ and $\kappa_3$ across layers, with mathematical prompts exhibiting lower values than Pile-CC. The higher-order cumulants $\kappa_6$ and $\kappa_7$ reveal a qualitative shift in the middle layers: mathematical prompts show higher values than Pile-CC in these layers. Together, these statistical signatures quantify how the model employs fundamentally different mechanisms for mathematical versus general text processing.

\section{Conclusion}

We propose cumulants as a probe for examining emergent statistical properties within LLMs and demonstrate their effectiveness in revealing representational patterns across model layers. Our new observables provide a mathematical framework for explicitly showing how neural networks learn higher-order correlations within data. Our experiments reveal that in later LLM layers, higher-order cumulants increase and vary significantly for structured prompts but much less for shuffled prompts. We also show during training, the model progressively learn higher-order correlations within data during training. Future research could explore the correlation between loss and cumulants, as well as the causal effects of directly manipulating cumulants.

\section{Reproducibility}
All the results contained in this work are reproducible by means of a GitHub repository that can be found at \href{https://github.com/karthikviswanathn/cumulant-expansion-llm}{https://github.com/karthikviswanathn/cumulant-expansion-llm}.
\section*{Acknowledgements}
We thank Boris Post for helpful discussions on the cumulant expansion of softmax entropy and Matteo Biagetti for the application of this framework to large language models.  We acknowledge the Dutch Research Council (NWO) in The
Netherlands for awarding this project access to the LUMI supercomputer, owned by the EuroHPC Joint Undertaking,
hosted by CSC (Finland) and the LUMI consortium through the ‘Computing Time on National Computer Facilities’
call. We used LUMI to run the computations required for this project.
\bibliography{sample}
\newpage
\clearpage
\appendix

\section{Derivations}
\label{sec:derivations}

We use the definition of the mean, the softmax function, and the KL-divergence. 

\begin{align*}
\langle S(\mathbf{X}) \rangle &= \frac{1}{N} \sum_{i = 1}^N S(X_i) \\
&= - \frac{1}{N} \sum_{i = 1}^N \sum_{\alpha = 1}^{\mathcal{V}} p_{i \alpha} \log p_{i \alpha} \nonumber \\
&=  - \frac{1}{N} \sum_{i = 1}^N \sum_{\alpha = 1}^{\mathcal{V}} p_{i \alpha} \log p_{i \alpha} - p_{i \alpha} \log p_{\mu \alpha}  - \frac{1}{N} \sum_{i = 1}^N \sum_{\alpha = 1}^{\mathcal{V}}  p_{i \alpha} \log p_{\mu \alpha} \nonumber \\
&= -  \sum_{\alpha = 1}^{\mathcal{V}} \log p_{\mu \alpha} \left(\frac{1}{N} \sum_{i = 1}^N   p_{i \alpha}\right) - \frac{1}{N} \sum_{i = 1}^ND(p_i||p_\mu) \nonumber \\
&= -  \sum_{\alpha = 1}^{\mathcal{V}} p_{\mu \alpha} \log p_{\mu \alpha} - \frac{1}{N} \sum_{i = 1}^ND(p_i||p_\mu)\\
&= S(\boldsymbol{\mu}) - \frac{1}{N} \sum_{i = 1}^ND(p_i||p_\mu)  
\end{align*}

Therefore we can obtain Eq.~\ref{eq:softmaxaroundmean}

\begin{equation}
\tag{\ref{eq:softmaxaroundmean}}
\boxed{
    \langle S(\mathbf{X}) \rangle = S(\boldsymbol{\mu}) - \frac{1}{N} \sum_{i = 1}^ND(p_i||p_\mu)
}
\end{equation}

Now, using the definition of the KL-divergence, and plugging in the distribution of the distribution of tokens from the softmax function, where the "partition function" $Z_\beta(\mathbf{X}) = \sum_{\alpha=1}^\mathcal{V} e^{\beta x_\alpha}$, and the definition of the cumulant generating function,

\begin{align*}
D_{\text{KL}}(p_\beta(X) \| p_\beta(\boldsymbol{\mu})) 
&= \sum_{\alpha=1}^\mathcal{V} \frac{e^{\beta x_\alpha}}{Z_\beta(\mathbf{X})} 
\log \frac{\frac{e^{\beta x_\alpha}}{Z_\beta(\mathbf{X})}}{\frac{e^{\beta \mu_\alpha}}{Z_\beta(\boldsymbol{\mu})}} \nonumber \\ 
&= \sum_{\alpha=1}^\mathcal{V} \frac{e^{\beta x_\alpha}}{Z_\beta(\mathbf{X})} \left(\beta (x_\alpha - \mu_\alpha) + \log \frac{Z_\beta(\boldsymbol{\mu})}{Z_\beta(\mathbf{X})}\right) \nonumber \\
&= \sum_{\alpha=1}^\mathcal{V} \frac{e^{\beta x_\alpha}}{Z_\beta(\mathbf{X})} \beta \delta x_\alpha + \log \sum_{\alpha=1}^\mathcal{V} \frac{e^{\beta x_\alpha}}{Z_\beta(\mathbf{X})} e^{-\beta \delta x_\alpha} \nonumber \\ 
&= \mathbb{E}_{p_\beta(X)} \left[\beta \delta x \right] + \log \mathbb{E}_{p_\beta(X)} \left[ e^{-\beta \delta x}\right]  \\
&= \sum_{n = 2}^{\infty} \frac{\beta^n}{n!} \kappa^{p_\beta(\mathbf{X})}_n(-\delta x)\quad (\because \text{definition of cumulant generating function} )
\end{align*}

Therefore we can obtain Eq.~\ref{eq:kldivergenceascumulants}

\begin{equation} 
\tag{\ref{eq:kldivergenceascumulants}}
\boxed{
    D_{\text{KL}}(p_\beta(X) \| p_\beta(\boldsymbol{\mu})) = \sum_{n = 2}^{\infty} \frac{\beta^n}{n!} \kappa^{p_\beta(\mathbf{X})}_n(-\delta X)
}
\end{equation}

Now, we use several properties of cumulants to express the deviation of the mean softmax entropy from entropy of the center as a function of cumulants of the sum of the logits' deviation from the center. 

\begin{align*}
\langle S(\mathbf{X}) \rangle 
&= S(\boldsymbol{\mu}) - \frac{1}{N} \sum_{i = 1}^N D(p_i||p_\mu) \\
&= S(\boldsymbol{\mu}) - \frac{1}{N} \sum_{i = 1}^N \sum_{n = 2}^{\infty} \frac{\beta^n}{n!} \kappa^{p_\beta(X_i)}_n(-\delta X_i) \nonumber \\ 
&= S(\boldsymbol{\mu}) -  \sum_{n = 2}^{\infty} \frac{\beta^n}{n!} \langle \kappa^{p_\beta(X_i)}_n(-\delta X_i)\rangle \\
&=  S(\boldsymbol{\mu}) - \frac{1}{N} \sum_{n = 2}^{\infty} \sum_{i = 1}^N  \frac{\beta^n}{n!} \kappa^{p_\beta(X_i)}_n(-\delta X_i) \nonumber \\  
&= S(\boldsymbol{\mu}) - \frac{1}{N} \sum_{n = 2}^{\infty}   \frac{\beta^n}{n!} \kappa^{p_\beta(\mathbf{X})}_n\left(-\sum_{i = 1}^N \delta X_i \right) \quad (\because \text{additive property of cumulants})
\end{align*}

Therefore we obtain Eq.~\ref{eq:finalrelation}

\begin{equation}
\tag{\ref{eq:finalrelation}}
\boxed{
     \langle S(\mathbf{X}) \rangle = S(\boldsymbol{\mu}) - \frac{1}{N} \sum_{n = 2}^{\infty}   \frac{\beta^n}{n!} \kappa^{p_\beta(\mathbf{X})}_n\left(-\sum_{i = 1}^N \delta X_i \right)
}
\end{equation}

\clearpage
\section{Supplementary Figures}
\label{sec:suppfigs}

\subsection{Cumulants for the prompts in the Pile dataset, on GPT2-Large}
\label{subsec:otherprompts}

\begin{figure}[t]
    \centering
    \includegraphics[width=1.0\textwidth]{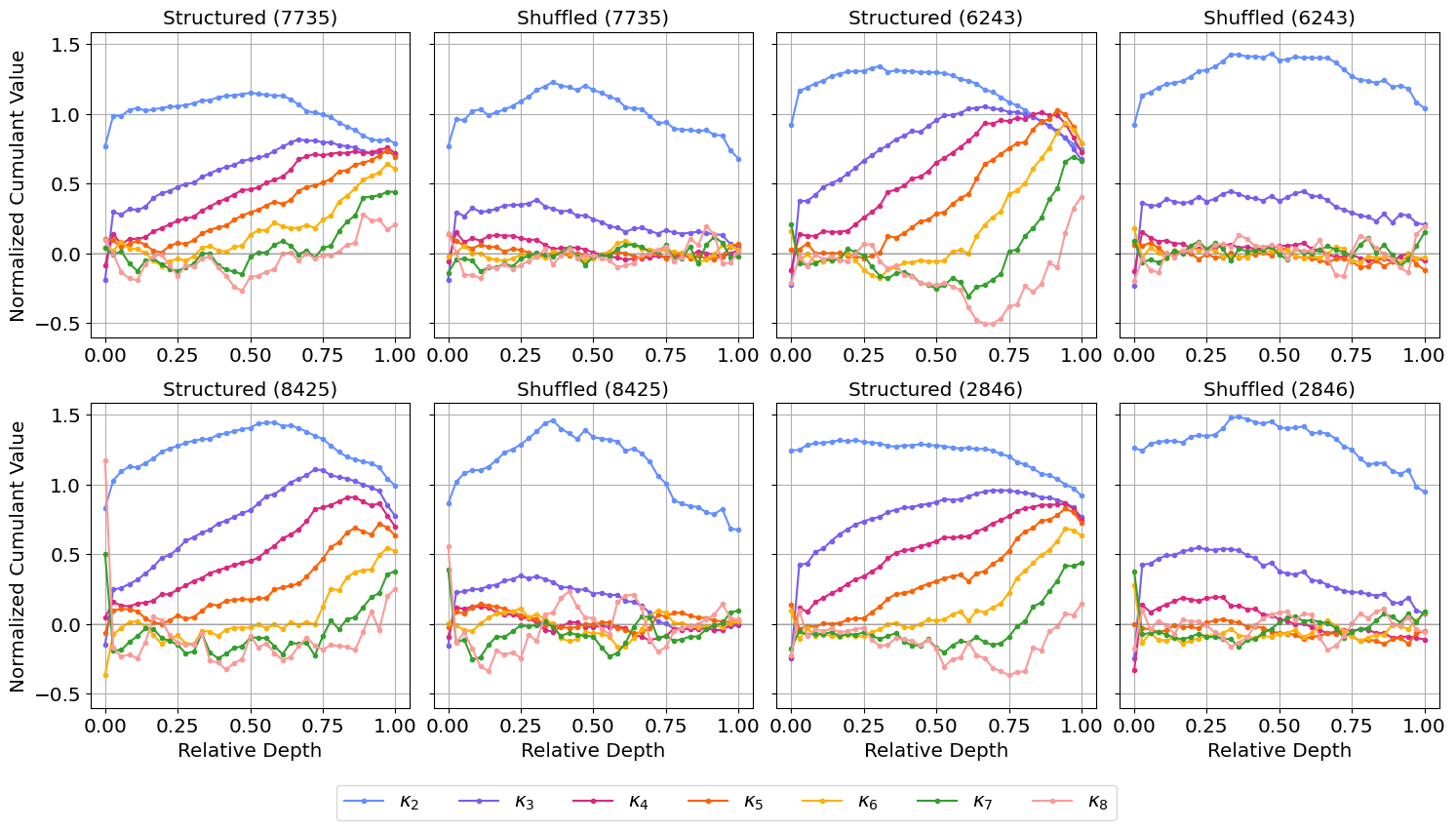}
    \caption{{\bfseries Cumulants Across Prompts.}  
    Cumulants of structured and shuffled versions of four randomly selected prompts from the Pile-10k dataset~\cite{NeelNanda_pile-10k}. 
    Each panel corresponds to a different prompt, and the numbers on the axis title represent the prompt number in the Pile-10K dataset.
    Structured prompts consistently show richer and more structured cumulant profiles compared to their shuffled counterparts.}
    \label{fig:manyprompts}
\end{figure}

\begin{figure}[h]
    \centering
    \includegraphics[width=1.0\textwidth]{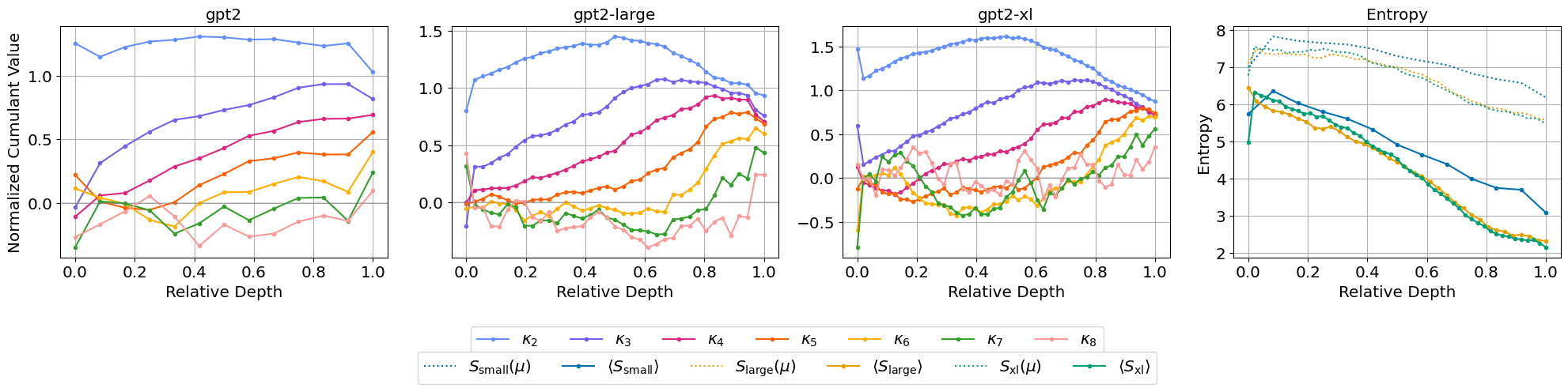}
    \caption{{\bfseries Cumulants and Entropy Across Model Sizes.} 
    Left three panels: cumulants of structured prompts across layers for three models from the GPT-2 family of increasing size (small, large, xl). 
    Right panel: comparing the mean softmax entropy and the softmax entropy of the center across the same models. 
    The structured prompts exhibit more pronounced cumulant variation and a larger entropy gap, consistent across model size.}
    \label{fig:differenetsizemodels}
\end{figure}

Fig.4 show the cumulants of structured and shuffled prompts for four different randomly selected prompts within the Pile-10k dataset~\cite{NeelNanda_pile-10k}, and Fig.5 show the result of a structured and shuffled prompt for different size models from the GPT2 family.

\subsection{Verification of Cumulants}
\label{subsec:verificationcumulants}

In this section we verify that the cumulants of $\delta X = \sum_{i=1}^n \delta X_i$ are indeed equal to the sum of the per‐token cumulants $\delta X_i$ following the discussion in Section \ref{sec: cumulants_as_observables}. We fix a prompt and focus on a single layer $L$. For each of $N$ Monte Carlo trials, we:

\begin{enumerate}
  \item For each token $i$, sample a vocabulary index $t_i$ according to the $p_\beta(X_i)$.
  \item Calculate how "far" $X_i$ is from the center along the direction $t_i$, by measuring $\delta X_{i, t_i} = X_{i,t_i} - \mu_{t_i}$. 
  \item  Sum over tokens to obtain a single realization of the aggregate deviation:  $\delta X = \sum_i \delta X_{i, t_i}$
\end{enumerate}

\begin{figure}
\centering
\includegraphics[width=0.45\textwidth]{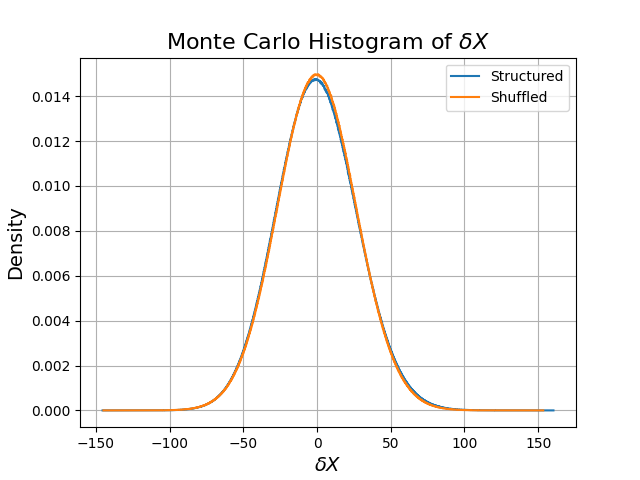}
\caption{{\bfseries Histogram of $\boldsymbol{\delta X}$.} We use $N = 2 \times 10^8$ Monte Carlo simulations for prompt 3218 at layer 20 of GPT2-Large for the structured (blue) and the shuffled (orange) prompts.}
\label{fig:montecarlo}
\end{figure}

We plot the histogram of $\delta X$ obtained from $N = 2 \times 10^8$ Monte Carlo simulations for both the structured and shuffled versions of prompt 3218 from the Pile-10K dataset, using the GPT2-Large model at layer 20 (Figure~\ref{fig:montecarlo}). We compare the cumulant values from the two methods in Table \ref{tab:cumulant_comparison_grid}.

\begin{table}[h]
    \centering
    \vspace{0.5em}
    \begin{tabular}{|l|c|c|c|c|c|c|}
        \hline
        \textbf{Type} & \multicolumn{2}{c|}{\textbf{2nd Cumulant}} & \multicolumn{2}{c|}{\textbf{3rd Cumulant}} & \multicolumn{2}{c|}{\textbf{4th Cumulant}} \\
        \hline
        & Monte Carlo & Average & Monte Carlo & Average & Monte Carlo & Average \\
        \hline
        Structured & 1.432 & 1.432 & 0.986 & 0.981 & 0.623 & 0.616 \\
        \hline
        Shuffled   & 1.351 & 1.430 & 0.238 & 0.294 & $-0.015$ & $-0.128$ \\
        \hline
    \end{tabular}
    \caption{{\bfseries Monte Carlo vs. Token-Wise Averaged Cumulants.}  
    Comparison of cumulant estimates obtained via Monte Carlo sampling and averaging cumulants over tokens for structured and shuffled prompts. The columns correspond to the (2nd, 3rd, 4th) cumulants.}
    \label{tab:cumulant_comparison_grid}
\end{table}

\section{Cumulant calculation}
\label{sec:cumulantcalculation}

\begin{algorithm2e}[h]
\caption{Layer‐wise Cumulant Computation for a Text Prompt}
\label{alg:cumulants_all_layers}
\SetAlgoLined
\KwIn{Language Model $f$ with $L$ layers, prompt $P$ with $T$ tokens, max cumulant order $K$.}
\KwOut{Cumulants $\kappa_n^{(\ell)}$ for $n=1,\dots,K$ and $\ell=1,\dots,L$.}

Tokenize $P$ into $w_{1:T}$ and run forward pass of $f$ on $w_{1:T}$\;
Extract for each layer $\ell=1,\dots,L$ the logits from TunedLens $X_t^{(\ell)}\in\mathbb R^V$ at each position $t$\;

\For{$\ell\leftarrow1$ \KwTo $L$}{
  \For{$t\leftarrow1$ \KwTo $T$}{
    $p_t^{(\ell)} \leftarrow \mathrm{softmax}\bigl(X_t^{(\ell)}\bigr)$\;
    
  }
  
  $\displaystyle p_\mu^{(\ell)} \leftarrow \frac{1}{T}\sum_{t=1}^T p_t^{(\ell)}$ \tcp*{Compute center distribution $p_\mu$}\
  $\displaystyle \mu^{(\ell)} \leftarrow \log p_\mu^{(\ell)}$;\

  \For{$t\leftarrow1$ \KwTo $T$}{
    $\displaystyle \delta X_t^{(\ell)}\leftarrow X_t^{(\ell)}\;-\;\mu^{(\ell)}$\;
    }

  \tcp{Compute token‐wise moments and cumulants}
  \For{$t\leftarrow1$ \KwTo $T$}{
    \For{$n\leftarrow1$ \KwTo $K$}{
      $m_n^{(\ell)}(t)\leftarrow\sum_{v=1}^V p_t^{(\ell)}(v)\,\bigl[\delta X_t^{(\ell)}(v)\bigr]^n$\;
      \tcp{Obtaining cumulants from moments where $B_{n,k}$ are incomplete Bell polynomials.}
      $\kappa_n^{(\ell)}(t)=\sum_{k=1}^n(-1)^{k-1}(k-1)!B_{n, k}\left(m_1^{(\ell)}(t), \ldots, m_n^{(\ell)}(t)\right)$\;
    }
  }

  \tcp{Average cumulants across tokens}
  \For{$n\leftarrow1$ \KwTo $K$}{
    $\displaystyle \kappa_n^{(\ell)}\leftarrow \frac{1}{T}\sum_{t=1}^T \kappa_n^{(\ell)}(t)$\;
  }
}
\Return $\{\kappa_n^{(\ell)}\}_{\substack{\ell=1..L\\n=1..K}}$\;
\end{algorithm2e}

\end{document}